\pdfoutput=1
\documentclass[11pt]{article}
\usepackage{acl}
\usepackage{times}
\usepackage{latexsym}
\usepackage[T1]{fontenc}
\usepackage[utf8]{inputenc}
\usepackage{microtype}
\title{Discourse over Discourse: The Need for an Expanded Pragmatic Focus in Conversational AI}

\author{S.M. Seals \\
  Air Force Research Laboratory \\
  Oak Ridge Institute for Science and Education \\
  Wright State University \\
  \texttt{s.m.seals@outlook.com} \\\And
  Valerie L. Shalin \\
  Wright State University \\
  }
  
\renewenvironment{quote}{
\noindent     
\topsep=0pt   
\parskip=0pt  
\leftskip=0pt 
\rightskip=0pt 
\parfillskip=0pt 

\itshape  
\list{}{
\leftmargin 0em   
\rightmargin 0em  
}
\item
}
{\endlist}

\begin{document}
\maketitle
\begin{abstract}
The summarization of conversation, that is, discourse over discourse, elevates pragmatic considerations as a pervasive limitation of both summarization and other applications of contemporary conversational AI. Building on impressive progress in both semantics and syntax, pragmatics concerns meaning in the practical sense. In this paper, we discuss several challenges in both summarization of conversations and other conversational AI applications, drawing on relevant theoretical work. We illustrate the importance of pragmatics with so-called star sentences, syntactically acceptable propositions that are pragmatically inappropriate in conversation or its summary. Because the baseline for quality of AI is indistinguishability from human behavior, we draw heavily on the psycho-linguistics literature, and label our complaints as "Turing Test Triggers" (TTTs). We discuss implications for the design and evaluation of conversation summarization methods and conversational AI applications like voice assistants and chatbots.
\end{abstract}

\section{Introduction}
\footnote{The views expressed are those of the author and do not necessarily reflect the official policy or position of the Department of the Air Force, the Department of Defense, or the U.S. government.}
The summarization of conversation, a case of discourse over a discourse, clearly illustrates a series of pragmatic limitations in contemporary conversational AI applications. While there has been some previous work examining pragmatic issues in conversational AI (i.e., \cite{bao-etal-2022-learning, kim-etal-2020-will, kim-etal-2021-perspective, Nath2020, wu-ong-2021-pragmatically}), additional progress depends on understanding the source of limitations in current applications. We aim to contribute to both theory and applications by examining several recurrent pragmatic limitations associated with both conversation summarization models and other conversational AI applications.

No doubt, applications of conversational NLP have achieved considerable performance improvements and are an area of increasing research focus. Deep learning has enabled the development of personal assistants \cite{radford_language_2019}, ASR \cite{hinton_deep_2012} and machine translation \cite{sutskever_sequence_2014}, as well as large language models that can generate seemingly natural sentences \cite{brown_language_2020, devlin-etal-2019-bert, zhang_opt_2022}. This work has spurred the development of models that can employ certain characteristically human aspects of dialogue \cite{kim-etal-2020-will, majumder-etal-2020-mime, wu-etal-2019-proactive}. Likewise, automatic summarization has expanded from sentence compression and document summarization \cite{bhandari-etal-2020-evaluating, li-etal-2017-cascaded, nayeem-etal-2018-abstractive, wang-etal-2020-heterogeneous} to summaries of casual conversations \cite{chen-yang-2021-structure, goo_abstractive_2018} and meetings \cite{gillick_global_2009}. Such summaries are intended to provide an account of 'what the exchange was about'. Moreover, performance of these models has improved, prompting empirical work on new evaluation metrics \cite{bhandari-etal-2020-evaluating}. Despite these performance gains, there are remaining areas for improvement in both conversational summarization and conversational AI more broadly. We illustrate the remaining challenges in this area with ill-conceived examples inspired by conversational AI systems \cite{gratch_distress_2014}, conversation summarization models, \cite{Manas2021} and author interactions with chatbots and voice assistants. Like Chomsky's star sentences, these examples have clear pragmatic deficiencies that trigger the Turing Test criterion. No competent speaker would construct such discourse.

Our approach helps structure user frustrations with conversational AI systems documented in the HCI literature, typically highlighting complaints about conversational skills separately from other usability concerns \cite{brandtzaeg_why_2017, folstad_chatbots_2019,liao_what_2016, luger_like_2016, porcheron_voice_2018, zamora_im_2017}. In so doing, we promote synergy between applied and basic research endeavors that address language in use. 

In particular, we suggest below that pragmatic limitations (and user frustrations) of current conversational AI systems are addressed by the pragmatic theory of \emph{relevance}. Next, we articulate two sub-themes for understanding and addressing these limitations: preservation of meaning and incorporation of external context. Throughout, we inventory a series of pragmatic failures in different domains, and discuss them in the context of relevant theoretical work in linguistics, psycho-linguistics, and cognitive psychology. 

\section{User Relevance}
Conversations (and comprehensive summaries) preserve \emph{relevance}. When people engage in conversation, they expect that their conversational partners will make contributions that are relevant to the ongoing dialogue and consistent with the accepted purpose of the conversation. When users interact with conversational AI applications, they have similar expectations for those systems (i.e., \cite{zamora_im_2017}). \cite{Grice1975} initially proposed that the expectation of relevance is due to a \emph{cooperative principle} and that the expectation of related utterances is due to a maxim of \emph{relation}. 

\cite{sperber_relevance_1986} revised this explanation. Instead of an explicit cooperative principle, they proposed that the search for relevance is a basic feature of human cognition \cite{Wilson2013}. A given input is relevant if processing that input generates a worthwhile difference to a recipient's representation of the world \cite{Wilson2013}. Relevance is necessarily determined in context. For a given stimulus to be relevant in the prevailing context, it must be worth the recipient's processing effort and be the most relevant stimulus available that is compatible with the person's abilities and preferences \cite{Wilson2013}. Given that all other factors are equal, if one stimulus requires less processing effort than another, that stimulus will be the most relevant.

Relevance theory emphasizes that any content a conversational AI system provides that is difficult to read, missing important information, or is incorrect reduces relevance to the end user. When users encounter this type of content, they must either expend additional effort to understand the information or search elsewhere for more appropriate information. Consider a source conversation and its attempted automated  \emph{star} summary below. 

\begin{quote}
Interviewer: \emph{Can you give me an example of that?}

Patient: \emph{If somebody ... annoys me I'll probably let them know they're annoying me until they stop.}

(TTT) Patient: \emph{Yeah} 

Pragmatically-appropriate summary: \emph{I would let them know they're annoying me.}
\end{quote}

The hypothetical summary of the participant response, \emph{yeah}, is not be acceptable either in the source conversation or as a summary. From a theoretical perspective, such a response is not pragmatically appropriate. The interviewer's question is an example of an indirect speech act \cite{Clark1979} with both literal and indirect meaning.  

An answer that addresses just the literal form of the question, as in the example, violates the expectation of relevance. The intended, indirect meaning of the question 
requires a response that addresses that meaning. The direct meaning response, either in actual conversation or it's summary, creates the impression of a flippant patient who flouts relevance expectations. Indeed, \cite{Clark1979} demonstrated that responses to indirect speech acts predominantly answer the indirect meaning, as in the above actual response \cite{Clark1979}. From a practical perspective, the suggested summary fails to convey the patient's actual response.

A summarization method that does not preserve the original meaning of the conversation risks a false conclusion \cite{Wilson2013}. Should a user recognize that this answer is likely misleading (because it violates expected responses of indirect speech acts), the user must revisit the original conversation to determine the correct account. This failure of relevance owing to the associated additional effort fatally reduces the utility of the summarization model.

The example above, and related empirical work in HCI, illustrates two fundamental and related dimensions of relevance; meaning and contextual awareness. Next, we examine these two dimensions more closely and pragmatic failures that can arise from each.

\section{Meaning and Inference}

In order to generate relevant content, conversational AI systems must respond to the meaning of a user's utterance, with respect to its contribution to the conversation as a whole. In this section, we discuss some pragmatic aspects of meaning as they relate to conversational AI. 

Meaning is often reduced to the domain of classical semantics in philosophy. Accordingly, discourse meaning lies in propositional content. \emph{Propositions} are the purported elementary units of meaning and represent truth measurable in the world \cite{Levinson}. The flawed summary above omitted its explicit propositional content. From a psychological perspective, propositions bridge the distinction between specific words and their corresponding concepts \cite{buschke_memory_1979, forster_visual_1970, goetz_representation_1981, graesser_structural_1980, kintsch_representation_1974, kintsch_reading_1973}. 

One of Chomksy's most important insights is that sentences often contain multiple propositions with complex interrelationships and dependencies(i.e., \cite{forster_visual_1970, graesser_structural_1980}). The multiple sentences of discourse exacerbate the problem of recovering interrelationships, otherwise known as coherence. Automated responses that are unable to account for between sentence relationships will compromise coherence \cite{mcnamara_are_1996}. The resulting additional effort impairs comprehension and challenges relevance. \cite{beck_revising_1991}. The voice assistant example below illustrates this point.

\begin{quote}
    User: \emph{I want to go to Cleveland, is there any construction that would slow down my trip?}
    
    (TTT) Assistant: \emph{Getting directions to Cleveland [does not provide traffic information].}
    
    Pragmatically-appropriate assistant: \emph{Traveling to Cleveland will take 3 hours. There are currently no traffic delays. [Pulls up directions]}
\end{quote}

In a typical automated reply, the assistant only addresses the first proposition in the user's utterance, wanting directions to a given city. The second proposition about traffic delays is ignored. Pragmatically acceptable responses require representation of both propositions and a knowledge-driven inference---the user is more generally concerned about travel time to Cleveland in which construction from traffic is merely an instance of the concern. From this perspective, an appropriate summary of the exchange is not 'the user wanted directions to Cleveland', but rather 'the user was concerned about potential travel delays on a road trip to Cleveland.' 

\section{Context}
Context has two dimensions. The first concerns factors that are internal to the conversation. This dimension includes aspects of context that can be understood given the prior content of the conversation. The second deals with factors external to the specific lexical content of the conversation, typically features of its physical setting. 
Psychologists invoke the constructs of memory to explain context-related processing.  One of these constructs concerns semantic memory, which contains general knowledge that enables the kind of inference just described.  A second construct concerns episodic memory for specific events \cite{tulving_episodic_1972} and includes associated aspects such as who, when and where \cite{nyberg_general_1996}. Semantic and episodic memory influences both conversational and external context.  

\subsection{Conversational Context}

The specific topics, words and phrases that members of the conversation use determine a conversational context that participates in the comprehension of subsequent utterances.  When humans process discourse, they expand a representation of the unfolding content. As comprehension proceeds, propositions combine into \emph{macropropositions} that recursively combine to form the \emph{macrostructure} \cite{VanDijk1983}. Speakers use and update these representations throughout conversation to understand their partners utterances and make relevant follow-up contributions \cite{clark_common_1983, Isaacs1987, Lockridge2002}.

Furthermore, users want systems to exploit the kinds of information that are plausibly in conversational context. In fact, \cite{folstad_chatbots_2019} found that users cared more about such conversational abilities of chatbots than personalities or appearance. Users would like systems that can ask clarifying questions and remember previous interactions with that user, particularly if the interactions occur close together in time \cite{luger_like_2016}. 

Users would like to understand how systems work \cite{liao_what_2016, zamora_im_2017}, what kinds of tasks they can do \cite{liao_what_2016, luger_like_2016, zamora_im_2017}, and when systems acquire new capabilities \cite{luger_like_2016}. The ideal collaborative system is aware of the user's status and intentions and responds accordingly \cite{liao_what_2016} with personalized recommendations or help users consider multiple options \cite{folstad_chatbots_2019}. Pragmatic failures arise when conversational AI systems lack the corresponding conceptual representations that a human partner would maintain during discourse. In the following subsections, we discuss some of these pragmatic failures.

\subsubsection{Memory for Dialogue Topics}

Speakers in conversation create expectations about what pieces of information are shared between all members of the conversation. This representation is called \emph{common ground} \cite{Clark1996}. Common ground represents each individual's beliefs about what information their partner knows, based on community membership and past personal experience, including the experience of the ongoing exchange. Speakers develop models of what information their partners have processed and update them as the conversation progresses \cite{pickering_toward_2004}.Conversational AI applications generally do not create these kinds of representations (i.e., \cite{rollo_carpenterexistor_cleverbot_2022}) or do so in a limited fashion. 

To generate pragmatically appropriate content, conversational AI applications need memory for topics that were previously addressed in the conversation, represented in \emph{personal common ground}. This type of memory is episodic with respect to the conversation-- it requires awareness of what topics were discussed, the answers that were provided, and who said what. When systems do not represent the content of prior conversation, they generate pragmatically inappropriate content, as we demonstrate next. 

\begin{quote}
    (TTT) Interviewer: \emph{Do you still go to therapy now?}
    
    Patient: \emph{I don't, I never went to therapy.}
\end{quote}

In using the word \emph{still}, the automated interviewer presupposes previous therapy, without having established that fact in the conversation. A more appropriate question here would be to first ask if the patient has been to therapy and then ask if the patient still attends therapy, conditional on an affirmative response to the first question. Moreover, this illustrates that comprehensive summaries are predicated on coherent dialogue. A comprehensive summary of this exchange requires both information about if the patient has previously been to therapy and if the patient is currently in therapy. Given such an exchange, a proper summary mirrors the familiar phrasing of clinical notes: the patient is not currently in therapy and denies having been to therapy in the past.

\subsubsection{Propose Relevant Next Turns}

Similarly, systems must propose new turns that are consistent with prior conversational content. This is particularly apparent for applications that involve multiple turns of dialogue, but is evident for single turn interactions where a user repeatedly interacts with a system as well. The example below is from a mental health chatbot for practicing CBT skills. The chatbot asks the user to provide a thought and then proposes follow-up questions that are intended to help the user process the thought and make changes to their perspective. 

\begin{quote}
    Bot: \emph{Does this thought assume others are thinking negatively about you?}
    
    User: \emph{No}
    
    Bot: \emph{Does your thought place too much blame on others?}
    
    User: \emph{No it doesn't}
    
    Bot: \emph{Does it place too much blame on you?}
    
    User: \emph{Nope}
    
    (TTT) Bot: \emph{Can you rewrite the thought without any distortions?}
    
    User: \emph{It doesn't have any}

    (TTT) Bot: \emph{How did you do?}
\end{quote}

This chatbot does not account for the prior conversational context in generating the question \emph{can you rewrite the thought without any distortions?}. A more appropriate implementation here might propose a different strategy for the user. This failure to account for prior conversation is further illustrated by the last question, which does not respond directly to the content of the previous utterance. A more appropriate utterance would account for the user's failure to answer the previous question and propose an alternative course of action. Moreover, an appropriate summary of this exchange would highlight that failure. 

\subsubsection{Correction of Prior Content}

Systems need the ability to correct and update content that was previously introduced in dialogue. Information is introduced into common ground via a collaborative process between partners \cite{clark_collaborating_1987, brennan_conceptual_1996} and partners may make changes or updates until all members are satisfied. Moreover, speakers routinely engage in self-correction to revise prior mistakes \cite{Schegloff1977}. Conversational AI systems have notable difficulties with these situations despite their pervasiveness in dialogue, as illustrated below.

\begin{quote}
    User: \emph{I want to book a ticket to Newark, sorry New York}
    
    (TTT) Bot: \emph{There are multiple destinations, please select from the list below. [list includes Newark, NJ, Newark DE, and Newark NJ.]}
\end{quote}

The chatbot response does not handle the user's attempted self-correction. There are several possible answers that would be more appropriate here. First, the system could suggest New York, as the user intended. Second, the system could clarify if Newark or New York was the intended destination. This option is not exactly what the user requested, but would be a better response than only showing options for Newark. Previous methods that have been proposed for clarification generation (i.e., \cite{benotti-blackburn-2021-recipe, kim_deciding_2021, majumder-etal-2021-ask}) could be employed to address this issue. Similarly, a pragmatically sensitive summary of this exchange would be \emph{The user wanted to book a ticket to New York.}. It is unnecessary to capture the misstatement or its correction.

\subsubsection{Semantic Imprecision}

Anaphora is a common class of semantically imprecise words. To understand the intended meaning, people (or conversational AI systems) must determine the correct referent. This is conventionally achieved via linear order; given a pronoun, the correct referent is the most proximal noun \cite{carden_backwards_1982}. However, examination of discourse in use demonstrates this approach is flawed-- pronouns regularly precede their referents or refer to remotely mentioned subjects \cite{carden_backwards_1982}. \cite{van_hoek_backwards_1997} proposed that people use imaginary perspectives called \emph{conceptual reference points} to view discourse and assign connections between items \cite{langacker_reference-point_1993}. Use of these reference points allows people to assign pronouns to distal or upcoming referents.  

In NLP, pronouns are often ignored (as in the case of removal with stop word lists \cite{Nothman2019}) or addressed via replacement. While some work has found replacement to work well for discourse in a constrained environment (i.e., \cite{chen-yang-2021-structure}), obvious errors can result when replacement is applied to naturalistic discourse, as illustrated in the example conversation summary below. 

\begin{quote}
Interviewer: \emph{What do you do when you're annoyed?}

Patient: \emph{[provides response]}

Interviewer: \emph{Can you give me an example of that?}

Patient: \emph{Uh, if someone ... annoys me I would let them know they're annoying me until they stop.}

(TTT) Summary: \emph{What do you do when \textbf{they} are annoying?}
\end{quote}

The pragmatic failure of this example is subtle, but has clear implications for reader understanding. The interviewer asks the patient about their activities, specifically, what the patient does when \emph{the patient} is annoyed. The patient provides a requested example, but uses the pronoun \emph{they} to refer to a hypothetical annoying third party. The summarization model proposes a summary for the interviewer's question that replaces the second pronoun \emph{you} with the pronoun \emph{they}. 

While the proposed summary is grammatically correct, it has two pragmatic failures. First, the summary does not include a referent for the pronoun. The meaning of \emph{you} can be determined from the context-- the interviewer is clearly speaking with a patient. The referent for \emph{they} is unclear. The pragmatically appropriate summary in this case would be the original question.

Pronouns also convey point of view information \cite{van_hoek_pronominal_2010}. For systems that require a complete description of the speaker's experience, incorrect pronoun assignment risks incorrect conclusions that lead to pragmatic failures. In the example above, the proposed summary changes the focus of the conversation. The interviewer's question was intended to gather information about the patient's behavior. The summary question shifts the focus of attention to the patient being annoyed.

Ambiguous pronouns violate the expected structure of summaries and therefore influence the coherence of reader's mental representations of text \cite{Bransford1972, Bransford1973a}. Specifically, ambiguous pronouns reduce the coherence of reader's representations of discourse content \cite{mcnamara_are_1996}. 

Speakers use point of view to account for their partners' perspectives \cite{Lockridge2002}, prior expertise with a topic, \cite{Isaacs1987}, and perceptual abilities \cite{clark_common_1983}. 

\subsection{External Context} 
When speakers engage in conversation with others, they expect that their conversation partners are aware of salient features of their shared external context \cite{Clark1981}. This includes both the physical environment and relevant background knowledge. Shared context controls detail in the exchange, allowing for the elimination of the obvious, and explicit emphasis on the non-obvious. 

Previous research on user expectations for chatbots and conversational agents has found that users expect systems to account for external context and find it frustrating when they are unable to do so \cite{liao_what_2016}. Users want systems to be aware of their status and intentions and respond accordingly \cite{liao_what_2016}. Users create expectations of information that systems should know and want systems to use that information \cite{luger_like_2016}. Including external context in the design of conversational AI applications, surely a challenging goal,  will produce systems that are more consistent with users' expectations and more straightforward to use.

Empirical research has demonstrated that humans regularly utilize external context in conversation to provide the right amount of detail \cite{van_der_henst_testing_2004}, account for a partner's expertise \cite{Isaacs1987}, and create references that partners will understand \cite{clark_common_1983}. Where previous work has incorporated context, the focus has largely been on conversational context. We suggest that this approach is insufficient-- even if conversational context is represented perfectly, pragmatic failures will arise from a lack of appropriate awareness of relevant aspects of external context. In the following subsections, we discuss some of these pragmatic failures and relevant theoretical work. 

\subsubsection{Episodic Features}

Speakers regularly use words and expressions like \emph{today} that are semantically imprecise and are understood with reference to the current context \cite{Levinson}. These language functions are examined in the theoretical area of \emph{deixis} \cite{Levinson}. They are tolerated in conversation when they are efficient to articulate and interpret. A previously un-grounded \emph{there} for example, becomes tolerable when the speaker glances at the intended referent. For NLP applications, these expressions pose a semantic interpretation problem due to an impoverished representation of episodic conversation features.

Consistent with the notion of common ground, \cite{barwise_situations_1983} proposed that utterances require interpretation with respect to three situations. The \emph{discourse situation} represents facts that someone might observe about the conversation. The \emph{resource situation} includes the relationships between the speakers and facts known to all members. The \emph{described situation} includes facts that could verify or falsify an utterance. This taxonomy reveals the scope of contextual features that are not always incorporated in the development of conversational AI applications.

Conversational AI applications often fail to incorporate not only external facts about the physical setting, 
like the user's location (represented in the \emph{discourse situation}), but also semantic knowledge that all members of the conversation already know (represented in the \emph{resource situation}). Examples of background knowledge include conceptual knowledge \cite{speer_conceptnet_2017}, domain specific knowledge \cite{gaur__2018}, attribute information \cite{zhang_collaborative_2016}, commonsense knowledge \cite{davis_commonsense_2015} and or information about the user.

This lack of appropriate awareness of external context generates several issues for conversational AI systems. First, relevance theory emphasizes that systems that lack awareness of relevant external context are unlikely to generate optimally relevant content for users. The voice assistant example below illustrates a pragmatic failure arising from a lack of this type of knowledge.

\begin{quote}
    User: \emph{Is there a heat warning today?}
    
    (TTT) Assistant: \emph{I found this on the web} [Provides news article about heat wave in the UK when the user is in the US.]
    
    Pragmatically-appropriate assistant: \emph{Yes, there is a heat warning effect in [area] until [time].}
\end{quote}

As is typical of conventional conversation, the user does not specify their easily inferred location and the voice assistant fails to account for the user's location. The user does not receive the expected answer and must search elsewhere. A more appropriate answer is illustrated in the second response- it provides relevant information tailored to the user making the request. Moreover, a proper summary would actually \emph{add} inferred content:  the user wanted to know of heat warnings in [area].  Indeed, users create expectations about the kinds of information that conversational AI systems should have (i.e., location information from their profile) and would like systems to make use of that information \cite{luger_like_2016}.

\subsubsection{Conceptual Knowledge}
Similarly, pragmatic failures can arise from an interaction between lack of episodic awareness and a lack of requisite semantic knowledge. The following voice assistant example illustrates this point.

\begin{quote}
    User: \emph{Do I need gloves today?}
    
    (TTT) Assistant: \emph{Not much sun in the forecast today.}
    
    Pragmatically-appropriate assistant: \emph{The temperature is X degrees today.}
\end{quote}

In order to answer this question correctly, two types of knowledge are required. The first is knowledge of the current situation. The system must have an awareness of what day it is, where the user is located, and what the weather forecast is for that day in that location. Second, the system needs the conceptual knowledge about what gloves are and why the user might want to wear them (i.e., because it might be cold outside). This type of knowledge is often discussed under common sense reasoning \cite{davis_commonsense_2015}. The sample response successfully demonstrates awareness of the current situation. However, the system clearly lacks the requisite conceptual knowledge and gives an irrelevant answer about the amount of sun in the weather forecast. A more appropriate answer might include the low temperature for the day. A proper summary would also add content:  The user asked if it was cold enough for gloves in [area] on [date]. For the user to determine if gloves are required, they must search elsewhere for the temperature or ask a more specific question. Indeed, users want systems to understand such intentions and respond accordingly \cite{liao_what_2016}.

\subsubsection{Default Reasoning}

Default reasoning addresses situations where available information is incomplete and conclusions need to be made based on what is generally true \citep{brewka_introduction_2012}. Conversation often contains instances that require default reasoning. Conversational AI applications need the ability to handle these instances. Consider the case of traffic delays illustrated in the example below. 

\begin{quote}
    User: \emph{I want to go to Cleveland, are there any traffic delays?}
    
    (TTT) Assistant: \emph{Getting directions to Cleveland [does not provide information about delays].}
    
    Pragmatically-appropriate assistant: \emph{Traveling to Cleveland will take 3 hours. There are no current delays. [Pulls up directions]}
\end{quote}

A pragmatically sensitive response to the query would acknowledge all sources of traffic delays likely to impact the user's trip. While the prototypical example of a traffic delay is construction, a pragmatically sensitive response would also account for other possible delays, such as a high probability snow storm. Similarly, a pragmatically sensitive summary of the exchange is: the user was concerned about travel delays from [area] to Cleveland and the system provided information about possible delays.

Default reasoning is a type of nonmonotonic reasoning. Unlike traditional logic, nonmonotonic reasoning addresses situations where new information can invalidate old conclusions \cite{brewka_introduction_2012}. In the above example, a pragmatically sensitive answer would account for the likelihood of a specific delay. It would not be pragmatically sensitive to provide a warning about a possible delay from a minor traffic slowdown several hours ahead.

\subsubsection{Inconsistent Details}
Another type of nonmonotonic reasoning that presents a challenge for conversational AI is reasoning given inconsistent details \cite{brewka_introduction_2012}. When two pieces of information are inconsistent, reasoners must determine what parts of the available information should be disregarded and what parts should be retained. Humans are generally able to resolve these sorts of inconsistencies \cite{johnson-laird_reasoning_2004}. Conversational AI applications that lack these abilities will generate pragmatic errors, as illustrated in the voice assistant example below:

\begin{quote}
    User: \emph{Remind me on Friday August 4th at 5:00 to order groceries. [Friday is August 5th, not August 4th]}
    
    (TTT) Assistant: \emph{Done [creates reminder for Thursday August 4th at 5:00]}
    
    Pragmatically-appropriate assistant: \emph{Did you mean Thursday August 4th or Friday August 5th?}
\end{quote}

To determine the correct action, the assistant needs to detect and then resolve the inconsistency of which piece of information the user intended, \emph{Friday} or \emph{the 4th}. 

A more appropriate response here would be to request clarification, as in the sample appropriate response. Failure to detect and resolve the inconsistent results in what is known as conversational breakdown \cite{ashktorab_resilient_2019}. A comparable summary would report the corrected date. Inconsistency is compounded where simultaneous activity occurs with conversation (such as in meetings). Comprehensive summaries in this area require reasoning about the state of the world based on dialogue.

To effectively resolve these inconsistencies, systems need the ability to detect  inconsistent information and intervene. Previous work has developed methods for proposing clarification questions when conversational AI systems are unsure of the meaning of a user's utterance (i.e., \cite{benotti-blackburn-2021-recipe, kim_deciding_2021, majumder-etal-2021-ask}). Similar methods could be employed to address situations where users provide inconsistent information as in the example. Moreover, these methods could address situations where other information indicates inconsistency, such as a user who asks a voice assistant to create a new calendar event that would overlap an existing event. 

\subsubsection{Domain Specificity}

Lastly, external context pragmatic failures can arise from specific application environments where users have prior expertise with a given topic. These applications need communal common ground with the intended user \cite{Clark1996} to support appropriate audience design \cite{Bell1984}. For example, it is appropriate to define new anatomy terminology in lecture summaries or automated tutoring systems. This same terminology should not be defined in a summary of a meeting between doctors discussing the statuses of current patients. Similarly, virtual assistants, ASR systems, content filters, and other applications need an awareness of domain content when developed for domain-specific applications. One humorous real world example is a profanity filter for a virtual archaeology conference that banned the word \emph{bone} \cite{ferreira_profanity_2020}.

\section{Discussion}
We have demonstrated that the challenges to automated conversation summary, what we have termed discourse over discourse, are symptomatic of a more fundamental, general problem regarding the absence of attention to the role of pragmatics in automated conversation. We proposed that the pragmatic failures of conversational AI systems are captured by \emph{relevance theory} \cite{Wilson2013}. Relevance suggests two key issues for conversational AI systems: preservation of meaning and awareness of external context.

While previous NLP work has examined pragmatic issues separately, in language models \cite{pandia-etal-2021-pragmatic, ettinger-2020-bert, gubelmann-handschuh-2022-context, wang-etal-2021-calibrate-listeners}, downstream tasks \cite{nie-etal-2020-pragmatic, schuz-zarriess-2021-decoupling, zhang-etal-2022-improving}, dialogue systems and conversational models \cite{bao-etal-2022-learning, kim-etal-2020-will, kim-etal-2021-perspective, Nath2020, wu-ong-2021-pragmatically}, our integrative approach is intended to distill and taxonomize recurrent foundational themes to motivate a theoretical framework and coordinated research efforts. Similarly, we suggest that a theoretical framework will facilitate response to the large body of applied work on human expectations for conversational AI applications \cite{ashktorab_resilient_2019, liao_what_2016, luger_like_2016, zamora_im_2017}. We aim to provide such a framework by integrating these issues with theoretical and empirical work in pragmatics. 

\subsection{Limitations and Ethical Considerations}
 
This class of work has several important limitations and ethical concerns. First, this work inherits privacy concerns common to these types of applications. Many of the features represented in external context are not necessarily directly accessible via the semantic content in a conversation (i.e., user location). While some users want systems to use this information \cite{luger_like_2016}, others may not. Systems should clearly illustrate needed information, intended use and storage. Users should be able to easily and accessibly customize what information they share. Moreover, it is important to avoid creating systems that provide a sub-optimal user experience for users who do not want to share information. \cite{zuboff_age_2020} points out that some products are essentially not usable without agreeing to the product's data use policy. Creating systems that can pose clarification questions is one way of addressing this issue. If a user asks a question that requires location information they have not shared, the system could ask if there is a specific location the user would like a response for. 

Second, our position could be interpreted as endorsing the development of deep learning models with high monetary and energy costs \cite{strubell-etal-2019-energy}. We point out that many of the issues we raise could be addressed with approaches that utilize pre-existing external knowledge \cite{valiant_knowledge_2006}, such as knowledge graphs \cite{miller_wordnet_1995, speer_conceptnet_2017} or lexicons \cite{gaur__2018, sheth_semantic_2005}, that can reduce the need to acquire this information through deep learning. 

\section{Conclusion}
Several pragmatic challenges recur across current conversational AI applications. Drawing on relevant theoretical work in linguistics and psycho-linguistics, we examine each of these challenges in detail. We illustrate our points with examples that are syntactically correct, but have clear pragmatic deficiencies and integrate our observations with HCI research that has examined user expectations and frustrations surrounding current conversational AI applications. These results contribute to a better understanding of current pragmatic challenges and suggest areas for improvement. Two of these needs most salient in our review are better connection to general knowledge and the external environment. Contrary to the notion of summaries as simplified versions of discourse, we observe that comprehensive summaries often require adding content to clearly to relevant general knowledge and aspects of the external environment. Future work can examine possible approaches to developing systems that can address these challenges and better meet the pragmatic expectations of users.

\bibliography{anthology,zotero}

\begin{thebibliography}{87}
\expandafter\ifx\csname natexlab\endcsname\relax\def\natexlab#1{#1}\fi

\bibitem[{Ashktorab et~al.(2019)Ashktorab, Jain, Liao, and
  Weisz}]{ashktorab_resilient_2019}
Zahra Ashktorab, Mohit Jain, Q.~Vera Liao, and Justin~D. Weisz. 2019.
\newblock \href {https://doi.org/10.1145/3290605.3300484} {Resilient
  {Chatbots}: {Repair} {Strategy} {Preferences} for {Conversational}
  {Breakdowns}}.
\newblock In \emph{Proceedings of the 2019 {CHI} {Conference} on {Human}
  {Factors} in {Computing} {Systems}}, pages 1--12, Glasgow Scotland Uk. ACM.

\bibitem[{Bao et~al.(2022)Bao, Ghosh, and Chai}]{bao-etal-2022-learning}
Yuwei Bao, Sayan Ghosh, and Joyce Chai. 2022.
\newblock \href {https://aclanthology.org/2022.acl-long.202} {Learning to
  mediate disparities towards pragmatic communication}.
\newblock In \emph{Proceedings of the 60th Annual Meeting of the Association
  for Computational Linguistics (Volume 1: Long Papers)}, pages 2829--2842,
  Dublin, Ireland. Association for Computational Linguistics.

\bibitem[{Barwise and Perry(1983)}]{barwise_situations_1983}
Jon Barwise and John Perry. 1983.
\newblock \emph{Situations and {Attitudes}}.
\newblock MIT Press, Cambridge, MA.

\bibitem[{Beck et~al.(1991)Beck, McKeown, Sinatra, and
  Loxterman}]{beck_revising_1991}
Isabel~L. Beck, Margaret~G. McKeown, Gale~M. Sinatra, and Jane~A. Loxterman.
  1991.
\newblock \href {https://doi.org/10.2307/747763} {Revising {Social} {Studies}
  {Text} from a {Text}-{Processing} {Perspective}: {Evidence} of {Improved}
  {Comprehensibility}}.
\newblock \emph{Reading Research Quarterly}, 26(3):251--276.
\newblock Publisher: [Wiley, International Reading Association].

\bibitem[{Bell(1984)}]{Bell1984}
Allan Bell. 1984.
\newblock Language style as audience design.
\newblock \emph{Language in Society}, 13(2):145--204.

\bibitem[{Benotti and Blackburn(2021)}]{benotti-blackburn-2021-recipe}
Luciana Benotti and Patrick Blackburn. 2021.
\newblock \href {https://doi.org/10.18653/v1/2021.naacl-main.320} {A recipe for
  annotating grounded clarifications}.
\newblock In \emph{Proceedings of the 2021 Conference of the North American
  Chapter of the Association for Computational Linguistics: Human Language
  Technologies}, pages 4065--4077, Online. Association for Computational
  Linguistics.

\bibitem[{Bhandari et~al.(2020)Bhandari, Gour, Ashfaq, Liu, and
  Neubig}]{bhandari-etal-2020-evaluating}
Manik Bhandari, Pranav~Narayan Gour, Atabak Ashfaq, Pengfei Liu, and Graham
  Neubig. 2020.
\newblock \href {https://doi.org/10.18653/v1/2020.emnlp-main.751}
  {Re-evaluating evaluation in text summarization}.
\newblock In \emph{Proceedings of the 2020 Conference on Empirical Methods in
  Natural Language Processing (EMNLP)}, pages 9347--9359, Online. Association
  for Computational Linguistics.

\bibitem[{Brandtzaeg and Følstad(2017)}]{brandtzaeg_why_2017}
Petter~Bae Brandtzaeg and Asbjørn Følstad. 2017.
\newblock \href {https://doi.org/10.1007/978-3-319-70284-1_30} {Why {People}
  {Use} {Chatbots}}.
\newblock In \emph{Internet {Science}}, Lecture {Notes} in {Computer}
  {Science}, pages 377--392, Cham. Springer International Publishing.

\bibitem[{Bransford and Johnson(1972)}]{Bransford1972}
John~D. Bransford and Marcia~K. Johnson. 1972.
\newblock \href {https://doi.org/10.1016/S0022-5371(72)80006-9} {Contextual
  prerequisites for understanding: {Some} investigations of comprehension and
  recall}.
\newblock \emph{Journal of Verbal Learning and Verbal Behavior},
  11(6):717--726.

\bibitem[{Bransford and Johnson(1973)}]{Bransford1973a}
John~D Bransford and Marcia~K Johnson. 1973.
\newblock Considerations of some problems of comprehension.
\newblock In \emph{Visual {Information} {Processing}}, pages 383--438. Academic
  Press.

\bibitem[{Brennan and Clark(1996)}]{brennan_conceptual_1996}
Susan~E. Brennan and Herbert~H. Clark. 1996.
\newblock \href {https://doi.org/10.1037/0278-7393.22.6.1482} {Conceptual pacts
  and lexical choice in conversation}.
\newblock \emph{Journal of Experimental Psychology: Learning Memory and
  Cognition}, 22(6):1482--1493.

\bibitem[{Brewka(2012)}]{brewka_introduction_2012}
Gerhard Brewka. 2012.
\newblock Introduction.
\newblock In \emph{Nonmonotonic {Reasoning}: {Logical} {Foundations} of
  {Commonsense}}. Cambridge University Press, Cambridge, MA.

\bibitem[{Brown et~al.(2020)Brown, Mann, Ryder, Subbiah, Kaplan, Dhariwal,
  Neelakantan, Shyam, Sastry, Askell, Agarwal, Herbert-Voss, Krueger, Henighan,
  Child, Ramesh, Ziegler, Wu, Winter, Hesse, Chen, Sigler, Litwin, Gray, Chess,
  Clark, Berner, McCandlish, Radford, Sutskever, and
  Amodei}]{brown_language_2020}
Tom Brown, Benjamin Mann, Nick Ryder, Melanie Subbiah, Jared~D Kaplan, Prafulla
  Dhariwal, Arvind Neelakantan, Pranav Shyam, Girish Sastry, Amanda Askell,
  Sandhini Agarwal, Ariel Herbert-Voss, Gretchen Krueger, Tom Henighan, Rewon
  Child, Aditya Ramesh, Daniel Ziegler, Jeffrey Wu, Clemens Winter, Chris
  Hesse, Mark Chen, Eric Sigler, Mateusz Litwin, Scott Gray, Benjamin Chess,
  Jack Clark, Christopher Berner, Sam McCandlish, Alec Radford, Ilya Sutskever,
  and Dario Amodei. 2020.
\newblock \href
  {https://proceedings.neurips.cc/paper/2020/file/1457c0d6bfcb4967418bfb8ac142f64a-Paper.pdf}
  {Language models are few-shot learners}.
\newblock In \emph{Advances in neural information processing systems},
  volume~33, pages 1877--1901. Curran Associates, Inc.

\bibitem[{Buschke and Schaier(1979)}]{buschke_memory_1979}
Herman Buschke and Aron~H. Schaier. 1979.
\newblock \href {https://doi.org/10.1016/S0022-5371(79)90304-9} {Memory units,
  ideas, and propositions in semantic remembering}.
\newblock \emph{Journal of Verbal Learning and Verbal Behavior},
  18(5):549--563.

\bibitem[{Carden(1982)}]{carden_backwards_1982}
Guy Carden. 1982.
\newblock Backwards anaphora in discourse context.
\newblock \emph{Journal of Linguistics}, 18(2):361--387.

\bibitem[{Carpenter/Existor(2022)}]{rollo_carpenterexistor_cleverbot_2022}
Rollo Carpenter/Existor. 2022.
\newblock \href {https://www.cleverbot.com/} {Cleverbot}.

\bibitem[{Chen and Yang(2021)}]{chen-yang-2021-structure}
Jiaao Chen and Diyi Yang. 2021.
\newblock \href {https://doi.org/10.18653/v1/2021.naacl-main.109}
  {Structure-aware abstractive conversation summarization via discourse and
  action graphs}.
\newblock In \emph{Proceedings of the 2021 Conference of the North American
  Chapter of the Association for Computational Linguistics: Human Language
  Technologies}, pages 1380--1391, Online. Association for Computational
  Linguistics.

\bibitem[{Clark(1979)}]{Clark1979}
Herbert~H. Clark. 1979.
\newblock \href {https://doi.org/10.1016/0010-0285(79)90020-3} {Responding to
  indirect speech acts}.
\newblock \emph{Cognitive Psychology}, 11(4):430--477.

\bibitem[{Clark(1996)}]{Clark1996}
Herbert~H. Clark. 1996.
\newblock \emph{Using language}.
\newblock Cambridge University Press.

\bibitem[{Clark and Marshall(1981)}]{Clark1981}
Herbert~H Clark and Catherine~R Marshall. 1981.
\newblock Definite reference and mutual knowledge.
\newblock In A.K. Joshi, B.L. Webber, and I.A. Sag, editors, \emph{Elements of
  discourse understanding}, pages 10--63. Cambridge University Press,
  Cambridge.
\newblock ISSN: 0749596X.

\bibitem[{Clark and Schaefer(1987)}]{clark_collaborating_1987}
Herbert~H Clark and Edward~F Schaefer. 1987.
\newblock \href {https://doi.org/10.1080/01690968708406350} {Collaborating on
  contributions to conversations}.
\newblock \emph{Language and Cognitive Processes}, 2(1):19--41.

\bibitem[{Clark et~al.(1983)Clark, Schreuder, and Buttrick}]{clark_common_1983}
Herbert~H. Clark, Robert Schreuder, and Samuel Buttrick. 1983.
\newblock \href {https://doi.org/10.1016/S0022-5371(83)90189-5} {Common ground
  at the understanding of demonstrative reference}.
\newblock \emph{Journal of Verbal Learning and Verbal Behavior},
  22(2):245--258.

\bibitem[{Davis and Marcus(2015)}]{davis_commonsense_2015}
Ernest Davis and Gary Marcus. 2015.
\newblock \href {https://doi.org/10.1145/2701413} {Commonsense reasoning and
  commonsense knowledge in artificial intelligence}.
\newblock \emph{Communications of the ACM}, 58(9):92--103.

\bibitem[{Devlin et~al.(2019)Devlin, Chang, Lee, and
  Toutanova}]{devlin-etal-2019-bert}
Jacob Devlin, Ming-Wei Chang, Kenton Lee, and Kristina Toutanova. 2019.
\newblock \href {https://doi.org/10.18653/v1/N19-1423} {{BERT}: Pre-training of
  deep bidirectional transformers for language understanding}.
\newblock In \emph{Proceedings of the 2019 Conference of the North {A}merican
  Chapter of the Association for Computational Linguistics: Human Language
  Technologies, Volume 1 (Long and Short Papers)}, pages 4171--4186,
  Minneapolis, Minnesota. Association for Computational Linguistics.

\bibitem[{Ettinger(2020)}]{ettinger-2020-bert}
Allyson Ettinger. 2020.
\newblock \href {https://doi.org/10.1162/tacl_a_00298} {What {BERT} is not:
  Lessons from a new suite of psycholinguistic diagnostics for language
  models}.
\newblock \emph{Transactions of the Association for Computational Linguistics},
  8:34--48.

\bibitem[{Ferreira(2020)}]{ferreira_profanity_2020}
Becky Ferreira. 2020.
\newblock \href
  {https://www.vice.com/en/article/dyzamj/a-profanity-filter-banned-the-word-bone-at-a-paleontology-conference}
  {A {Profanity} {Filter} {Banned} the {Word} '{Bone}' at a {Paleontology}
  {Conference}}.

\bibitem[{Forster(1970)}]{forster_visual_1970}
Kenneth~I. Forster. 1970.
\newblock \href {https://doi.org/10.3758/BF03210208} {Visual perception of
  rapidly presented word sequences of varying complexity}.
\newblock \emph{Perception \& Psychophysics}, 8(4):215--221.

\bibitem[{Følstad and Skjuve(2019)}]{folstad_chatbots_2019}
Asbjørn Følstad and Marita Skjuve. 2019.
\newblock \href {https://doi.org/10.1145/3342775.3342784} {Chatbots for
  customer service: user experience and motivation}.
\newblock In \emph{Proceedings of the 1st {International} {Conference} on
  {Conversational} {User} {Interfaces}}, {CUI} '19, pages 1--9, New York, NY,
  USA. Association for Computing Machinery.

\bibitem[{Gaur et~al.(2021)Gaur, Aribandi, Kursuncu, Alambo, Shalin,
  Thirunarayan, Beich, Narasimhan, and Sheth}]{Manas2021}
Manas Gaur, Vamsi Aribandi, Ugur Kursuncu, Amanuel Alambo, Valerie~L. Shalin,
  Krishnaprasad Thirunarayan, Jonathan Beich, Meera Narasimhan, and Amit Sheth.
  2021.
\newblock \href {https://doi.org/10.2196/20865} {Knowledge-infused abstractive
  summarization of clinical diagnostic interviews: {Framework} development
  study}.
\newblock \emph{JMIR Mental Health}, 8(5):1--19.

\bibitem[{Gaur et~al.(2018)Gaur, Kursuncu, Alambo, Sheth, Daniulaityte,
  Thirunarayan, and Pathak}]{gaur__2018}
Manas Gaur, Ugur Kursuncu, Amanuel Alambo, Amit Sheth, Raminta Daniulaityte,
  Krishnaprasad Thirunarayan, and Jyotishman Pathak. 2018.
\newblock \href {https://doi.org/10.1145/3269206.3271732} {" {Let} {Me} {Tell}
  {You} {About} {Your} {Mental} {Health}!" {Contextualized} {Classification} of
  {Reddit} {Posts} to {DSM}-5 for {Web}-based {Intervention}}.
\newblock In \emph{Proceedings of the 27th {ACM} {International} {Conference}
  on {Information} and {Knowledge} {Management}}, pages 753--762.

\bibitem[{Gillick et~al.(2009)Gillick, Riedhammer, Favre, and
  Hakkani-Tur}]{gillick_global_2009}
Dan Gillick, Korbinian Riedhammer, Benoit Favre, and Dilek Hakkani-Tur. 2009.
\newblock \href {https://doi.org/10.1109/ICASSP.2009.4960697} {A global
  optimization framework for meeting summarization}.
\newblock In \emph{2009 {IEEE} {International} {Conference} on {Acoustics},
  {Speech} and {Signal} {Processing}}, pages 4769--4772.
\newblock ISSN: 2379-190X.

\bibitem[{Goetz et~al.(1981)Goetz, Anderson, and
  Schallert}]{goetz_representation_1981}
Ernest~T. Goetz, Richard~C. Anderson, and Diane~L. Schallert. 1981.
\newblock \href {https://doi.org/10.1016/S0022-5371(81)90506-5} {The
  representation of sentences in memory}.
\newblock \emph{Journal of Verbal Learning and Verbal Behavior},
  20(4):369--385.

\bibitem[{Goo and Chen(2018)}]{goo_abstractive_2018}
Chih-Wen Goo and Yun-Nung Chen. 2018.
\newblock \href {https://doi.org/10.1109/SLT.2018.8639531} {Abstractive
  {Dialogue} {Summarization} with {Sentence}-{Gated} {Modeling} {Optimized} by
  {Dialogue} {Acts}}.
\newblock In \emph{2018 {IEEE} {Spoken} {Language} {Technology} {Workshop}
  ({SLT})}, pages 735--742.

\bibitem[{Graesser et~al.(1980)Graesser, Hoffman, and
  Clark}]{graesser_structural_1980}
Arthur~C. Graesser, Nicholas~L. Hoffman, and Leslie~F. Clark. 1980.
\newblock \href {https://doi.org/10.1016/S0022-5371(80)90132-2} {Structural
  components of reading time}.
\newblock \emph{Journal of Verbal Learning and Verbal Behavior},
  19(2):135--151.

\bibitem[{Gratch et~al.(2014)Gratch, Artstein, Lucas, Stratou, Scherer,
  Nazarian, Wood, Boberg, DeVault, Marsella, Traum, Rizzo, and
  Morency}]{gratch_distress_2014}
Jonathan Gratch, Ron Artstein, Gale Lucas, Giota Stratou, Stefan Scherer,
  Angela Nazarian, Rachel Wood, Jill Boberg, David DeVault, Stacy Marsella,
  David Traum, Skip Rizzo, and Louis-Philippe Morency. 2014.
\newblock The {Distress} {Analysis} {Interview} {Corpus} of human and computer
  interviews.
\newblock In \emph{Proceedings of {LREC} 2014 {May}}, pages 3123--3128.

\bibitem[{Grice(1975)}]{Grice1975}
H.P. Grice. 1975.
\newblock Logic and conversation.
\newblock In Peter Cole and Jerry~L. Morgan, editors, \emph{Syntax and
  semantics 3: {Speech} acts}, pages 41--58. Academic Press, New York.

\bibitem[{Gubelmann and Handschuh(2022)}]{gubelmann-handschuh-2022-context}
Reto Gubelmann and Siegfried Handschuh. 2022.
\newblock \href {https://aclanthology.org/2022.acl-long.315} {Context matters:
  A pragmatic study of {PLM}s{'} negation understanding}.
\newblock In \emph{Proceedings of the 60th Annual Meeting of the Association
  for Computational Linguistics (Volume 1: Long Papers)}, pages 4602--4621,
  Dublin, Ireland. Association for Computational Linguistics.

\bibitem[{Hinton et~al.(2012)Hinton, Deng, Yu, Dahl, Mohamed, Jaitly, Senior,
  Vanhoucke, Nguyen, Sainath, and Kingsbury}]{hinton_deep_2012}
Geoffrey Hinton, Li~Deng, Dong Yu, George~E. Dahl, Abdel-rahman Mohamed,
  Navdeep Jaitly, Andrew Senior, Vincent Vanhoucke, Patrick Nguyen, Tara~N.
  Sainath, and Brian Kingsbury. 2012.
\newblock \href {https://doi.org/10.1109/MSP.2012.2205597} {Deep {Neural}
  {Networks} for {Acoustic} {Modeling} in {Speech} {Recognition}: {The}
  {Shared} {Views} of {Four} {Research} {Groups}}.
\newblock \emph{IEEE Signal Processing Magazine}, 29(6):82--97.

\bibitem[{Isaacs and Clark(1987)}]{Isaacs1987}
Ellen~A. Isaacs and Herbert~H. Clark. 1987.
\newblock \href {https://doi.org/10.1037/0096-3445.116.1.26} {References in
  {Conversation} {Between} {Experts} and {Novices}}.
\newblock \emph{Journal of Experimental Psychology: General}, 116(1):26--37.

\bibitem[{Johnson-laird et~al.(2004)Johnson-laird, Girotto, and
  Legrenzi}]{johnson-laird_reasoning_2004}
P.~N. Johnson-laird, Vittorio Girotto, and Paolo Legrenzi. 2004.
\newblock \href {https://doi.org/10.1037/0033-295X.111.3.640} {Reasoning from
  inconsistency to consistency}.
\newblock \emph{Psychological Review}, 111(3):640--661.

\bibitem[{Kim et~al.(2020)Kim, Kim, and Kim}]{kim-etal-2020-will}
Hyunwoo Kim, Byeongchang Kim, and Gunhee Kim. 2020.
\newblock \href {https://doi.org/10.18653/v1/2020.emnlp-main.65} {Will {I}
  sound like me? improving persona consistency in dialogues through pragmatic
  self-consciousness}.
\newblock In \emph{Proceedings of the 2020 Conference on Empirical Methods in
  Natural Language Processing (EMNLP)}, pages 904--916, Online. Association for
  Computational Linguistics.

\bibitem[{Kim et~al.(2021{\natexlab{a}})Kim, Kim, and
  Kim}]{kim-etal-2021-perspective}
Hyunwoo Kim, Byeongchang Kim, and Gunhee Kim. 2021{\natexlab{a}}.
\newblock \href {https://doi.org/10.18653/v1/2021.emnlp-main.170}
  {Perspective-taking and pragmatics for generating empathetic responses
  focused on emotion causes}.
\newblock In \emph{Proceedings of the 2021 Conference on Empirical Methods in
  Natural Language Processing}, pages 2227--2240, Online and Punta Cana,
  Dominican Republic. Association for Computational Linguistics.

\bibitem[{Kim et~al.(2021{\natexlab{b}})Kim, Wang, Lee, and
  Kim}]{kim_deciding_2021}
Joo-Kyung Kim, Guoyin Wang, Sungjin Lee, and Young-Bum Kim. 2021{\natexlab{b}}.
\newblock \href {https://doi.org/10.1109/ASRU51503.2021.9688265} {Deciding
  whether to ask clarifying questions in large-scale spoken language
  understanding}.
\newblock In \emph{2021 {IEEE} {Automatic} {Speech} {Recognition} and
  {Understanding} {Workshop} ({ASRU})}, pages 869--876, Cartagena, Colombia.
  IEEE.

\bibitem[{Kintsch(1974)}]{kintsch_representation_1974}
Walter Kintsch. 1974.
\newblock \emph{The representation of meaning in memory}.
\newblock The representation of meaning in memory. Lawrence Erlbaum, Oxford,
  England.
\newblock Pages: vii, 279.

\bibitem[{Kintsch and Keenan(1973)}]{kintsch_reading_1973}
Walter Kintsch and Janice Keenan. 1973.
\newblock \href {https://doi.org/10.1016/0010-0285(73)90036-4} {Reading rate
  and retention as a function of the number of propositions in the base
  structure of sentences}.
\newblock \emph{Cognitive Psychology}, 5(3):257--274.

\bibitem[{Langacker(1993)}]{langacker_reference-point_1993}
Ronald~W Langacker. 1993.
\newblock \href {https://doi.org/10.1515/cogl.1993.4.1.1} {Reference-point
  constructions}.
\newblock \emph{Cognitive Linguistics}, 4(1):1--38.

\bibitem[{Levinson(2011)}]{Levinson}
Stephen~C. Levinson. 2011.
\newblock Deixis.
\newblock In Laurence~R. Horn and Gregory Ward, editors, \emph{The handbook of
  pragmatics}, pages 97--121. Blackwell Publishing Ltd.

\bibitem[{Li et~al.(2017)Li, Lam, Bing, Guo, and Li}]{li-etal-2017-cascaded}
Piji Li, Wai Lam, Lidong Bing, Weiwei Guo, and Hang Li. 2017.
\newblock \href {https://doi.org/10.18653/v1/D17-1221} {Cascaded attention
  based unsupervised information distillation for compressive summarization}.
\newblock In \emph{Proceedings of the 2017 Conference on Empirical Methods in
  Natural Language Processing}, pages 2081--2090, Copenhagen, Denmark.
  Association for Computational Linguistics.

\bibitem[{Liao et~al.(2016)Liao, Davis, Geyer, Muller, and
  Shami}]{liao_what_2016}
Q.~Vera Liao, Matthew Davis, Werner Geyer, Michael Muller, and N.~Sadat Shami.
  2016.
\newblock \href {https://doi.org/10.1145/2901790.2901842} {What {Can} {You}
  {Do}?: {Studying} {Social}-{Agent} {Orientation} and {Agent} {Proactive}
  {Interactions} with an {Agent} for {Employees}}.
\newblock In \emph{Proceedings of the 2016 {ACM} {Conference} on {Designing}
  {Interactive} {Systems}}, pages 264--275, Brisbane QLD Australia. ACM.

\bibitem[{Lockridge and Brennan(2002)}]{Lockridge2002}
Calion~B Lockridge and Susan~E Brennan. 2002.
\newblock Addressees' needs influence speakers' early syntactic choices.
\newblock \emph{Psychonomic Bulletin and Review}, 9(3):550--557.

\bibitem[{Luger and Sellen(2016)}]{luger_like_2016}
Ewa Luger and Abigail Sellen. 2016.
\newblock \href {https://doi.org/10.1145/2858036.2858288} {"{Like} {Having} a
  {Really} {Bad} {PA}": {The} {Gulf} between {User} {Expectation} and
  {Experience} of {Conversational} {Agents}}.
\newblock In \emph{Proceedings of the 2016 {CHI} {Conference} on {Human}
  {Factors} in {Computing} {Systems}}, pages 5286--5297, San Jose California
  USA. ACM.

\bibitem[{Majumder et~al.(2021)Majumder, Rao, Galley, and
  McAuley}]{majumder-etal-2021-ask}
Bodhisattwa~Prasad Majumder, Sudha Rao, Michel Galley, and Julian McAuley.
  2021.
\newblock \href {https://doi.org/10.18653/v1/2021.naacl-main.340} {Ask what{'}s
  missing and what{'}s useful: Improving clarification question generation
  using global knowledge}.
\newblock In \emph{Proceedings of the 2021 Conference of the North American
  Chapter of the Association for Computational Linguistics: Human Language
  Technologies}, pages 4300--4312, Online. Association for Computational
  Linguistics.

\bibitem[{Majumder et~al.(2020)Majumder, Hong, Peng, Lu, Ghosal, Gelbukh,
  Mihalcea, and Poria}]{majumder-etal-2020-mime}
Navonil Majumder, Pengfei Hong, Shanshan Peng, Jiankun Lu, Deepanway Ghosal,
  Alexander Gelbukh, Rada Mihalcea, and Soujanya Poria. 2020.
\newblock \href {https://doi.org/10.18653/v1/2020.emnlp-main.721} {{MIME}:
  {MIM}icking emotions for empathetic response generation}.
\newblock In \emph{Proceedings of the 2020 Conference on Empirical Methods in
  Natural Language Processing (EMNLP)}, pages 8968--8979, Online. Association
  for Computational Linguistics.

\bibitem[{McNamara et~al.(1996)McNamara, Kintsch, Songer, and
  Kintsch}]{mcnamara_are_1996}
Danielle~S McNamara, Eileen Kintsch, Nancy~Butler Songer, and Walter Kintsch.
  1996.
\newblock Are good texts always better? {Interactions} of text coherence,
  background knowledge, and levels of understanding in learning from text.
\newblock \emph{Cognition and instruction}, 14(1):1--43.
\newblock Publisher: Taylor \& Francis.

\bibitem[{Miller(1995)}]{miller_wordnet_1995}
George~A. Miller. 1995.
\newblock {WordNet}: {A} lexical database for {English}.
\newblock \emph{Communications of the ACM}, 38(11):39--41.

\bibitem[{Nath(2020)}]{Nath2020}
Anindita Nath. 2020.
\newblock \href {https://doi.org/10.1145/3379336.3381490} {Towards naturally
  responsive spoken dialog systems by modelling pragmatic-prosody correlations
  of discourse markers}.
\newblock \emph{International Conference on Intelligent User Interfaces,
  Proceedings IUI}, pages 128--129.
\newblock ISBN: 9781450375139.

\bibitem[{Nayeem et~al.(2018)Nayeem, Fuad, and
  Chali}]{nayeem-etal-2018-abstractive}
Mir~Tafseer Nayeem, Tanvir~Ahmed Fuad, and Yllias Chali. 2018.
\newblock \href {https://aclanthology.org/C18-1102} {Abstractive unsupervised
  multi-document summarization using paraphrastic sentence fusion}.
\newblock In \emph{Proceedings of the 27th International Conference on
  Computational Linguistics}, pages 1191--1204, Santa Fe, New Mexico, USA.
  Association for Computational Linguistics.

\bibitem[{Nie et~al.(2020)Nie, Cohn-Gordon, and
  Potts}]{nie-etal-2020-pragmatic}
Allen Nie, Reuben Cohn-Gordon, and Christopher Potts. 2020.
\newblock \href {https://doi.org/10.18653/v1/2020.findings-emnlp.173}
  {Pragmatic issue-sensitive image captioning}.
\newblock In \emph{Findings of the Association for Computational Linguistics:
  EMNLP 2020}, pages 1924--1938, Online. Association for Computational
  Linguistics.

\bibitem[{Nothman et~al.(2019)Nothman, Qin, and Yurchak}]{Nothman2019}
Joel Nothman, Hanmin Qin, and Roman Yurchak. 2019.
\newblock \href {https://doi.org/10.18653/v1/w18-2502} {Stop {Word} {Lists} in
  {Free} {Open}-source {Software} {Packages}}.
\newblock In \emph{Proceedings of {Workshop} for {NLP} {Open} {Source}
  {Software}}, pages 7--12.

\bibitem[{Nyberg et~al.(1996)Nyberg, McIntosh, Cabeza, Habib, Houle, and
  Tulving}]{nyberg_general_1996}
L~Nyberg, A~R McIntosh, R~Cabeza, R~Habib, S~Houle, and E~Tulving. 1996.
\newblock \href {https://doi.org/10.1073/pnas.93.20.11280} {General and
  specific brain regions involved in encoding and retrieval of events: what,
  where, and when.}
\newblock \emph{Proceedings of the National Academy of Sciences},
  93(20):11280--11285.

\bibitem[{Pandia et~al.(2021)Pandia, Cong, and
  Ettinger}]{pandia-etal-2021-pragmatic}
Lalchand Pandia, Yan Cong, and Allyson Ettinger. 2021.
\newblock \href {https://doi.org/10.18653/v1/2021.conll-1.29} {Pragmatic
  competence of pre-trained language models through the lens of discourse
  connectives}.
\newblock In \emph{Proceedings of the 25th Conference on Computational Natural
  Language Learning}, pages 367--379, Online. Association for Computational
  Linguistics.

\bibitem[{Pickering and Garrod(2004)}]{pickering_toward_2004}
Martin~J. Pickering and Simon Garrod. 2004.
\newblock \href {https://doi.org/10.1017/S0140525X04000056} {Toward a
  mechanistic psychology of dialogue}.
\newblock \emph{Behavioral and Brain Sciences}, 27:169--226.

\bibitem[{Porcheron et~al.(2018)Porcheron, Fischer, Reeves, and
  Sharples}]{porcheron_voice_2018}
Martin Porcheron, Joel~E. Fischer, Stuart Reeves, and Sarah Sharples. 2018.
\newblock \href {https://doi.org/10.1145/3173574.3174214} {Voice {Interfaces}
  in {Everyday} {Life}}.
\newblock In \emph{Proceedings of the 2018 {CHI} {Conference} on {Human}
  {Factors} in {Computing} {Systems}}, pages 1--12, Montreal QC Canada. ACM.

\bibitem[{Radford et~al.(2019)Radford, Wu, Child, Luan, Amodei, and
  Sutskever}]{radford_language_2019}
Alec Radford, Jeffrey Wu, Rewon Child, David Luan, Dario Amodei, and Ilya
  Sutskever. 2019.
\newblock Language {Models} are {Unsupervised} {Multitask} {Learners}.

\bibitem[{Schegloff et~al.(1977)Schegloff, Jefferson, and
  Sacks}]{Schegloff1977}
Emanuel~A. Schegloff, Gail Jefferson, and Harvey Sacks. 1977.
\newblock \href {https://doi.org/10.2307/413107} {The preference for
  self-correction in the organization of repair in conversation}.
\newblock \emph{Language}, 53(2):361--382.

\bibitem[{Sch{\"u}z and Zarrie{\ss}(2021)}]{schuz-zarriess-2021-decoupling}
Simeon Sch{\"u}z and Sina Zarrie{\ss}. 2021.
\newblock \href {https://aclanthology.org/2021.reinact-1.7} {Decoupling
  pragmatics: Discriminative decoding for referring expression generation}.
\newblock In \emph{Proceedings of the Reasoning and Interaction Conference
  (ReInAct 2021)}, pages 47--52, Gothenburg, Sweden. Association for
  Computational Linguistics.

\bibitem[{Sheth et~al.(2005)Sheth, Aleman-Meza, Arpinar, Bertram, Warke,
  Ramakrishanan, Halaschek, Anyanwu, Avant, Arpinar, and
  {others}}]{sheth_semantic_2005}
Amit Sheth, Boanerges Aleman-Meza, I~Budak Arpinar, Clemens Bertram, Yashodhan
  Warke, Cartic Ramakrishanan, Chris Halaschek, Kemafar Anyanwu, David Avant,
  F~Sena Arpinar, and {others}. 2005.
\newblock Semantic association identification and knowledge discovery for
  national security applications.
\newblock \emph{Journal of Database Management (JDM)}, 16(1):33--53.
\newblock Publisher: IGI Global.

\bibitem[{Speer et~al.(2017)Speer, Chin, and Havasi}]{speer_conceptnet_2017}
Robyn Speer, Joshua Chin, and Catherine Havasi. 2017.
\newblock {ConceptNet} 5.5: {An} {Open} {Multilingual} {Graph} of {General}
  {Knowledge}.
\newblock In \emph{31st {AAAI} conference on artificial intelligence}, pages
  4444--4451.

\bibitem[{Sperber and Wilson(1986)}]{sperber_relevance_1986}
Dan Sperber and Deirdre Wilson. 1986.
\newblock \emph{Relevance: {Communication} and cognition}, volume 142.
\newblock Harvard University Press, Cambridge, Massachusetts.

\bibitem[{Strubell et~al.(2019)Strubell, Ganesh, and
  McCallum}]{strubell-etal-2019-energy}
Emma Strubell, Ananya Ganesh, and Andrew McCallum. 2019.
\newblock \href {https://doi.org/10.18653/v1/P19-1355} {Energy and policy
  considerations for deep learning in {NLP}}.
\newblock In \emph{Proceedings of the 57th Annual Meeting of the Association
  for Computational Linguistics}, pages 3645--3650, Florence, Italy.
  Association for Computational Linguistics.

\bibitem[{Sutskever et~al.(2014)Sutskever, Vinyals, and
  Le}]{sutskever_sequence_2014}
Ilya Sutskever, Oriol Vinyals, and Quoc~V Le. 2014.
\newblock Sequence to {Sequence} {Learning} with {Neural} {Networks}.
\newblock In \emph{Advances in {Neural} {Information} {Processing} {Systems}},
  volume~27.

\bibitem[{Tulving(1972)}]{tulving_episodic_1972}
Endel Tulving. 1972.
\newblock Episodic and semantic memory.
\newblock In \emph{Organization of memory}, pages xiii, 423--xiii, 423.
  Academic Press, Oxford, England.

\bibitem[{Valiant(2006)}]{valiant_knowledge_2006}
Leslie~G Valiant. 2006.
\newblock \href {www.aaai.org} {Knowledge infusion}.
\newblock \emph{AAAI}, 6.

\bibitem[{Van~der Henst and Sperber(2004)}]{van_der_henst_testing_2004}
Jean-Baptiste Van~der Henst and Dan Sperber. 2004.
\newblock \href {https://doi.org/10.1057/9780230524125_7} {Testing the
  {Cognitive} and {Communicative} {Principles} of {Relevance}}.
\newblock In Ira~A. Noveck and Dan Sperber, editors, \emph{Experimental
  {Pragmatics}}, Palgrave {Studies} in {Pragmatics}, {Language} and
  {Cognition}, pages 141--171. Palgrave Macmillan UK, London.

\bibitem[{van Dijk and Kintsch(1983)}]{VanDijk1983}
Teun~A. van Dijk and Walter Kintsch. 1983.
\newblock \href {https://doi.org/10.2307/415483} {\emph{Strategies of
  {Discourse} {Comprehension}}}.
\newblock Academic Press, New York.
\newblock ISSN: 00978507.

\bibitem[{Van~Hoek(1997)}]{van_hoek_backwards_1997}
Karen Van~Hoek. 1997.
\newblock Backwards anaphora as a constructional category.
\newblock \emph{Functions of Language}, 4(1):47--82.

\bibitem[{Van~Hoek(2010)}]{van_hoek_pronominal_2010}
Karen Van~Hoek. 2010.
\newblock \href {https://doi.org/10.1093/oxfordhb/9780199738632.013.0034}
  {Pronominal {Anaphora}}.
\newblock In Dirk Geeraerts and Hubert Cuyckens, editors, \emph{The {Oxford}
  {Handbook} of {Cognitive} {Linguistics}}, pages 1--25. Oxford University
  Press.

\bibitem[{Wang et~al.(2020)Wang, Liu, Zheng, Qiu, and
  Huang}]{wang-etal-2020-heterogeneous}
Danqing Wang, Pengfei Liu, Yining Zheng, Xipeng Qiu, and Xuanjing Huang. 2020.
\newblock \href {https://doi.org/10.18653/v1/2020.acl-main.553} {Heterogeneous
  graph neural networks for extractive document summarization}.
\newblock In \emph{Proceedings of the 58th Annual Meeting of the Association
  for Computational Linguistics}, pages 6209--6219, Online. Association for
  Computational Linguistics.

\bibitem[{Wang et~al.(2021)Wang, White, Mu, and
  Goodman}]{wang-etal-2021-calibrate-listeners}
Rose Wang, Julia White, Jesse Mu, and Noah Goodman. 2021.
\newblock \href {https://doi.org/10.18653/v1/2021.findings-emnlp.83} {Calibrate
  your listeners! robust communication-based training for pragmatic speakers}.
\newblock In \emph{Findings of the Association for Computational Linguistics:
  EMNLP 2021}, pages 977--984, Punta Cana, Dominican Republic. Association for
  Computational Linguistics.

\bibitem[{Wilson and Sperber(2013)}]{Wilson2013}
Deirdre Wilson and Dan Sperber. 2013.
\newblock \href {https://doi.org/10.4324/9780203206966-21} {Relevance theory}.
\newblock \emph{Routledge Companion to Philosophy of Language}, pages 163--176.
\newblock ISBN: 9781136594083.

\bibitem[{Wu et~al.(2019)Wu, Guo, Zhou, Wu, Zhang, Lian, and
  Wang}]{wu-etal-2019-proactive}
Wenquan Wu, Zhen Guo, Xiangyang Zhou, Hua Wu, Xiyuan Zhang, Rongzhong Lian, and
  Haifeng Wang. 2019.
\newblock \href {https://doi.org/10.18653/v1/P19-1369} {Proactive human-machine
  conversation with explicit conversation goal}.
\newblock In \emph{Proceedings of the 57th Annual Meeting of the Association
  for Computational Linguistics}, pages 3794--3804, Florence, Italy.
  Association for Computational Linguistics.

\bibitem[{Wu and Ong(2021)}]{wu-ong-2021-pragmatically}
Zhengxuan Wu and Desmond~C. Ong. 2021.
\newblock \href {https://aclanthology.org/2021.scil-1.53} {Pragmatically
  informative color generation by grounding contextual modifiers}.
\newblock In \emph{Proceedings of the Society for Computation in Linguistics
  2021}, pages 438--445, Online. Association for Computational Linguistics.

\bibitem[{Zamora(2017)}]{zamora_im_2017}
Jennifer Zamora. 2017.
\newblock \href {https://doi.org/10.1145/3125739.3125766} {I'm {Sorry}, {Dave},
  {I}'m {Afraid} {I} {Can}'t {Do} {That}: {Chatbot} {Perception} and
  {Expectations}}.
\newblock In \emph{Proceedings of the 5th {International} {Conference} on
  {Human} {Agent} {Interaction}}, {HAI} '17, pages 253--260, New York, NY, USA.
  Association for Computing Machinery.

\bibitem[{Zhang et~al.(2022{\natexlab{a}})Zhang, Zhou, Wan, Zheng, Chang, and
  Hsieh}]{zhang-etal-2022-improving}
Cenyuan Zhang, Xiang Zhou, Yixin Wan, Xiaoqing Zheng, Kai-Wei Chang, and
  Cho-Jui Hsieh. 2022{\natexlab{a}}.
\newblock \href {https://aclanthology.org/2022.findings-acl.284} {Improving the
  adversarial robustness of {NLP} models by information bottleneck}.
\newblock In \emph{Findings of the Association for Computational Linguistics:
  ACL 2022}, pages 3588--3598, Dublin, Ireland. Association for Computational
  Linguistics.

\bibitem[{Zhang et~al.(2016)Zhang, Yuan, Lian, Xie, and
  Ma}]{zhang_collaborative_2016}
Fuzheng Zhang, Nicholas~Jing Yuan, Defu Lian, Xing Xie, and Wei-Ying Ma. 2016.
\newblock \href {https://doi.org/10.1145/2939672.2939673} {Collaborative
  {Knowledge} {Base} {Embedding} for {Recommender} {Systems}}.
\newblock In \emph{Proceedings of the 22nd \{{ACM} {SIGKDD}\} {International}
  {Conference} on {Knowledge} {Discovery} and {Data} {Mining}}, pages 353--362,
  San Francisco California USA. ACM.

\bibitem[{Zhang et~al.(2022{\natexlab{b}})Zhang, Roller, Goyal, Artetxe, Chen,
  Chen, Dewan, Diab, Li, Lin, Mihaylov, Ott, Shleifer, Shuster, Simig, Koura,
  Sridhar, Wang, and Zettlemoyer}]{zhang_opt_2022}
Susan Zhang, Stephen Roller, Naman Goyal, Mikel Artetxe, Moya Chen, Shuohui
  Chen, Christopher Dewan, Mona Diab, Xian Li, Xi~Victoria Lin, Todor Mihaylov,
  Myle Ott, Sam Shleifer, Kurt Shuster, Daniel Simig, Punit~Singh Koura, Anjali
  Sridhar, Tianlu Wang, and Luke Zettlemoyer. 2022{\natexlab{b}}.
\newblock \href {http://arxiv.org/abs/2205.01068} {{OPT}: {Open} {Pre}-trained
  {Transformer} {Language} {Models}}.
\newblock ArXiv:2205.01068 [cs].

\bibitem[{Zuboff(2020)}]{zuboff_age_2020}
Shoshana Zuboff. 2020.
\newblock \emph{The {Age} of {Surveillance} {Capitalism}: {The} {Fight} for a
  {Human} {Future} at the {New} {Frontier} of {Power}}, 1st edition.
\newblock PublicAffairs.

\end{thebibliography}
\bibliographystyle{acl_natbib}
\appendix
\end{document}